\documentclass[lettersize,journal]{IEEEtran}
\usepackage{amsmath,amsfonts,amssymb}
\usepackage[noend]{algorithmic}
\usepackage{algorithm}
\usepackage{array}
\usepackage[caption=false,font=normalsize,labelfont=sf,textfont=sf]{subfig}
\usepackage{textcomp}
\usepackage{stfloats}
\usepackage{url}
\usepackage{verbatim}
\usepackage{graphicx}
\usepackage{cite}
\usepackage{booktabs} 
\usepackage{multirow}
\usepackage{xcolor}
\usepackage[super]{nth}
\usepackage{tikz}
\newcommand*\circled[1]{\tikz[baseline=(char.base)]{
            \node[shape=circle,draw,inner sep=0.2pt] (char) {#1};}}

\usepackage{soul}
\usepackage{enumitem}

\usepackage[switch]{lineno}
\usepackage{pifont}

\begin{document}


\title{PrimeSVT: An Automated Memory-aware Pruning Framework with Prioritized Compression Policy for Spiking Vision Transformers}

\author{Rachmad Vidya Wicaksana Putra,~\IEEEmembership{Member,~IEEE,} Achyuta Muthuvelan, Alberto Marchisio,~\IEEEmembership{Member,~IEEE,} and Muhammad Shafique,~\IEEEmembership{Senior Member,~IEEE} 
\thanks{Rachmad Vidya Wicaksana Putra, Achyuta Muthuvelan, and Alberto Marchisio are with eBRAIN Lab, Division of Engineering, New York University (NYU) Abu Dhabi, United Arab Emirates (UAE); (e-mail: rachmad.putra@nyu.edu, am12729@nyu.edu, and alberto.marchisio@nyu.edu). \\
Muhammad Shafique is the Director of eBRAIN Lab, Division of Engineering, New York University (NYU) Abu Dhabi, United Arab Emirates (UAE); 
(e-mail: muhammad.shafique@nyu.edu).}
}


\maketitle


\begin{abstract}
The large sizes of Spiking Vision Transformers (SViTs) still hinder their embedded implementation, highlighting the need for model compression.
State-of-the-art works compress SViT models through unstructured pruning, which needs specialized hardware accelerators for their specific sparsity patterns to maximize efficiency gains. 
Moreover, their manual approach requires a huge design time to find an appropriate pruning setting for each network, thus making this approach not scalable. 
To address this limitation, we propose \textbf{\textit{PrimeSVT}}, a novel framework that performs automated \textit{memory-aware structured pruning on pre-trained SViT models}, thereby maximizing their efficiency gains during inference amenable to widely-used computing architectures.
To achieve this, PrimeSVT first sorts the SViT layers based on their sizes (i.e., number of parameters), identifies the targeted pruning layers based on their robustness under different pruning rates, then leverages this order for compressing the model layer-by-layer sequentially from the largest one to the smallest one (i.e., so-called \textit{prioritized compression policy}), while considering the user-defined constraints (i.e., acceptable accuracy and memory saving). 
In each layer, PrimeSVT employs \textit{channel-wise filter pruning} based on their L2-norm values to structurally remove the non-significant weights. 
Experimental results show that PrimeSVT saves 26.68\% memory through automated single-shot pruning, while preserving accuracy within 3\% (70.3\% without fine-tuning and 72.9\% with fine-tuning) from the original unpruned SViT model (73.3\%), thus meeting the accuracy and memory constraints. 
These show that our PrimeSVT framework enables design automation for SViTs and their embedded implementation.  
\end{abstract}


\vspace{-0.1cm}
\section{Introduction}
\label{Sec_Intro} 
\vspace{-0.1cm}

\begin{figure}[t]
\centering
\includegraphics[width=\linewidth]{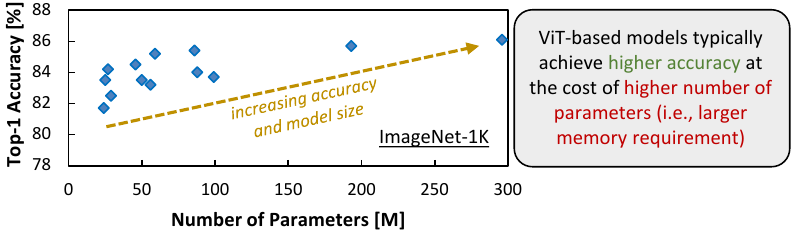}
\vspace{-0.7cm}
\caption{Advancements of ViT-based models successfully improve the accuracy, but also incur larger memory requirements (i.e., higher number of parameters); based on the data from~\cite{Ref_Han_SurveyViT_TPAMI22}. 
Note, M denotes millions [10$^6$] of parameters.}
\label{Fig_TrendViT}
\vspace{-0.3cm}
\end{figure}

Vision Transformers (ViTs) have rapidly emerged as state-of-the-art algorithms to solve diverse vision-based machine learning tasks, outperforming conventional artificial neural networks (ANNs)~\cite{Ref_Dosovitskiy_Transformers_ICLR21, touvron2021training, Ref_Khan_SurveyViT_CSUR22, Ref_Han_SurveyViT_TPAMI22}.
However, their large sizes and high power requirements make it difficult to realize their wide-scale embedded deployments. 
Current trends also indicate that larger ViTs typically offer higher accuracy than the smaller ones, as shown in Fig.~\ref{Fig_TrendViT}. 
Toward this, researchers recently leveraged Spiking Neural Network (SNN) operations for developing alternative low-power ViT models~\cite{Ref_Zhou_Spikformer_ICLR23, Ref_Yao_SpikeDrivenTransformer_NeurIPS23, Ref_Yao_SpikeDrivenTransformer2_ICLR24}, called \textit{Spiking Vision Transformers (SViTs)}.
However, their sizes remain too large for embedded deployments, underlining the need for model compression.

To compress SNN-based models, several methods have been proposed, such as pruning~\cite{Ref_Rathi_PruneQuantizeSNN_TCAD18}\cite{Ref_Putra_FSpiNN_TCAD20} and quantization~\cite{Ref_Rathi_PruneQuantizeSNN_TCAD18}\cite{Ref_Sorbaro_OptimSNN_FNINS20, Ref_Zou_MedianQuant_ISCAS20, Ref_Putra_QSpiNN_IJCNN21, Ref_Putra_QSViT_IJCNN25}.
Here, pruning is a prominent method, as it substantially removes insignificant paramaters (e.g., weights) while avoiding unacceptable accuracy drop. 
Although pruning method has been extensively explored by previous works in ANN domain~\cite{han2015deep, li2017pruning, marchisio2018prunet, zheng2022savit, yu2022width}, most of them require retraining to recover performance, thus requiring access to the complete (labeled) training data and incurring additional compute overheads.
Moreover, ANN-based pruning methods cannot be employed for SNN-based models, because of substantial differences between ANNs and SNNs in data representation and neuron operations.
Therefore, \textbf{the targeted research problem} in this paper is: \textit{how can we efficiently prune any given pre-trained SViT, while preserving its high accuracy and meeting the given memory constraint?} 
A solution to this problem promotes design automation for SViTs and their efficient embedded implementation.

\begin{table}[t]
\caption{The state-of-the-art pruning methods for SViTs.}
\vspace{-0.1cm}
\scriptsize
\centering
\begin{tabular}{|c|c|c|c|c|c|}
\hline
  \textbf{Work} & \begin{tabular}[c]{@{}c@{}} \textbf{Pruning} \\ \textbf{Target} \end{tabular}  & \begin{tabular}[c]{@{}c@{}} \textbf{Structured} \\ \textbf{Pruning} \end{tabular} & \begin{tabular}[c]{@{}c@{}} \textbf{Model} \\ \textbf{Compress.} \end{tabular} & \begin{tabular}[c]{@{}c@{}} \textbf{Large} \\ \textbf{Dataset} \end{tabular} & \begin{tabular}[c]{@{}c@{}} \textbf{Automated} \\ \textbf{Pruning} \end{tabular} \\ \hline \hline 
  \begin{tabular}[c]{@{}c@{}} STSFP~\cite{Ref_Zhou_STSFP_TCDS25} \end{tabular} & \begin{tabular}[c]{@{}c@{}} Token \end{tabular} & \ding{56} & \ding{56} & \ding{56} & \ding{56} \\  \hline
  \begin{tabular}[c]{@{}c@{}} STATA~\cite{Ref_Zhuge_TokenSparse_ICML24} \end{tabular} & \begin{tabular}[c]{@{}c@{}} Token \end{tabular} & \ding{56} & \ding{56} & \ding{52} & \ding{56} \\  \hline
  \begin{tabular}[c]{@{}c@{}} Sparse-\\spikformer\\ \cite{Ref_Liu_Sparsespikformer_ICASSP24} \end{tabular} & \begin{tabular}[c]{@{}c@{}} Token \& \\ Weights \end{tabular} & \ding{56} & \ding{52} & \ding{56} & \ding{56} \\  \hline
  \begin{tabular}[c]{@{}c@{}} \textbf{PrimeSVT}\\ \textbf{(ours)} \end{tabular} & Weights & \ding{52} & \ding{52} & \ding{52} & \ding{52} \\  \hline
\end{tabular}
\label{Tab_SOTA}
\vspace{-0.3cm}
\end{table}

\subsection{State-of-the-art of Pruning Methods for SViTs and Their Limitations}
\label{Sec_Intro_SOTA}

To perform SViT compression, several state-of-the-art works have proposed different pruning techniques. 
For instance, recent works employ the spatial-temporal spiking feature pruning (STSFP)~\cite{Ref_Zhou_STSFP_TCDS25} and the sparsification with timestep-wise anchor token and dual alignments (STATA)~\cite{Ref_Zhuge_TokenSparse_ICML24} to target input data/token sparsification, hence their model sizes are not reduced. 
Another work (i.e., Sparsespikformer~\cite{Ref_Liu_Sparsespikformer_ICASSP24}) prunes tokens and weights through a lottery ticket hypothesis.
\textit{Despite their benefits, these state-of-the-art have several limitations as summarized below (an overview in Table~\ref{Tab_SOTA}).}
\begin{itemize}[leftmargin=*]
    \item They use unstructured pruning, which needs specific hardware accelerators tailored to efficiently exploit the sparsity patterns and data indexing. 
    Otherwise, the efficiency gains are limited. 
    \item They also employ a manual pruning approach, which is not scalable for handling different possible SViT models as well as accuracy and memory constraints.
    \item STSFP and Sparsespikformer have not used complex and large datasets (e.g., ImageNet-1K~\cite{Ref_Deng_ImageNet_CVPR09}), which is necessary for benchmarking considering real-world environments.
\end{itemize}
These limitations indicate that \textit{an alternative pruning method is required to compress any given SViT models while meeting the accuracy and memory constraints}.

\subsection{Case Study and Related Research Challenges}
\label{Sec_Intro_Challenges}

\begin{figure}[t]
\centering
\includegraphics[width=\linewidth]{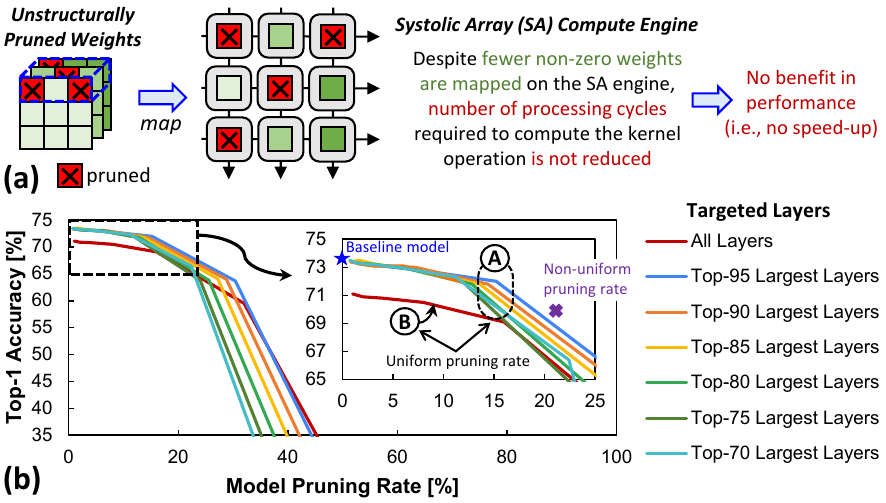}
\vspace{-0.6cm}
\caption{\textbf{(a)} Deploying unstructurally pruned weights on the widely-used systolic array (SA)-based compute engine does not lead to performance benefit (speed-up). 
\textbf{(b)} Accuracy profiles after implementing different structured pruning rates on different targeted layers of the state-of-the-art SDTv2 model~\cite{Ref_Yao_SpikeDrivenTransformer2_ICLR24} considering the ImageNet-1K~\cite{Ref_Deng_ImageNet_CVPR09} and different pruning scenarios.
Note, the accuracy of the baseline model (SDTv2) is obtained through our experiments based on the open-source codes from the original authors.}
\label{Fig_CaseStudy}
\vspace{-0.3cm}
\end{figure}

Unstructured pruning in the state-of-the-art typically results in irregularly distributed zero weights within filters, hence requiring specialized hardware accelerators to exploit the sparsity patterns and maximize the efficiency gains. 
For instance, the widely-used systolic array (SA) compute engine~\cite{Ref_Basu_SNNicSurvey_CICC22, Ref_Jouppi_TPU_ISCA17} expects dense weight mapping and neural operations. 
Therefore, irregularly distributed zero weights are still mapped in a regular fashion on the array, which results in no performance improvement (no speed-up); see Fig.~\ref{Fig_CaseStudy}(a). 
This observation indicates that the structured pruning is more hardware-friendly and practical than the unstructured one, hence it should be considered for SViT compression.
However, employing structured pruning on pre-trained SViTs is also a non-trivial task.

To illustrate the challenges of structured pruning on the pre-trained SViT models, we conduct an experimental case study\footnote{Details of the experimental setup is provided in Sec.~\ref{Sec_EvalMethod}} with the following settings.
\begin{itemize}[leftmargin=*]
    \item We perform the channel-wise pruning\footnote{Details of the channel-wise pruning is discussed in Sec.~\ref{Sec_PrimeSVT_TargetLayers_PruneAlg}} on the weight filters from the targeted layers of the SViT model. 
    \item \textit{Scenario-1 ``uniform pruning rate''}: 
    We investigate the impact of applying the same pruning rate uniformly across all targeted layers.
    Afterward, we also vary the pruning rates and the targeted layers (e.g., all layers, top-95\% largest layers, etc.). 
    \item \textit{Scenario-2 ``non-uniform pruning rate''}: 
    We manually set different pruning rates for different targeted layers to achieve competitive accuracy and memory saving.
    Hence, the final model pruning rate is the ratio between the total number of pruned weights and original unpruned ones. 
    \item We consider the state-of-the-art Spike-driven Transformer v2 (SDTv2)~\cite{Ref_Yao_SpikeDrivenTransformer2_ICLR24} and the ImageNet-1K dataset~\cite{Ref_Deng_ImageNet_CVPR09}.
\end{itemize}
\textit{Experimental results are shown in Fig.~\ref{Fig_CaseStudy}(b), from which we draw the following observations.}
\begin{itemize}[leftmargin=*]
    \item In general, pruning across fewer targeted layers reduces the possibility of achieving higher model reduction and high accuracy, since it is more likely to discard important weights and cause significant accuracy degradation; see~\circled{A}.
    \item Different layers may have different sensitivity levels under pruning. Therefore, pruning sensitive layers may lead to a notable accuracy drop despite considering a larger number of targeted layers (or even all layers); see~\circled{B}.  
    \item Non-uniform pruning rate has the potential to provide higher accuracy with higher pruning rate than the uniform pruning. 
\end{itemize}
\textit{Based on these observations, we identify the research challenges below.}
\begin{itemize}[leftmargin=*]
    \item Pruning process should carefully consider targeted layers within the SViT model to minimize the negative impact of sensitive layers on accuracy.
    \item Non-uniform pruning rate approach should be employed to achieve good trade-off between accuracy and memory saving.
    \item Pruning process should be able to meet different possible accuracy and memory constraints.
\end{itemize}

\vspace{-0.2cm}
\subsection{Our Novel Contributions}
\label{Sec_Intro_Novelty}

\begin{figure}[t]
\centering
\includegraphics[width=\linewidth]{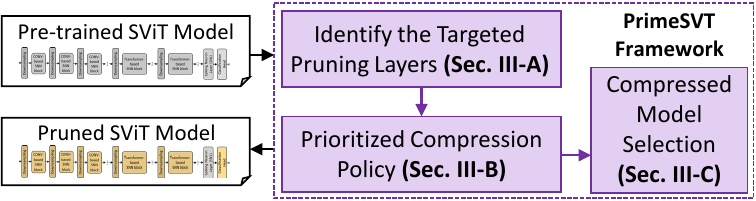}
\vspace{-0.6cm}
\caption{Overview of our novel contributions in this work.}
\label{Fig_Novelty}
\vspace{-0.3cm}
\end{figure}

To address the targeted problem and related challenges, we propose \textbf{\textit{PrimeSVT}}, \textit{a novel automated memory-aware \underline{Pr}uning framework with pr\underline{i}oritized co\underline{m}pr\underline{e}ssion policy for \underline{SV}i\underline{T} models to meet the accuracy and memory constraints}. 
It is also the first work that offers automated single-shot structured pruning for SViT compression.
It employs the following key ideas (an overview is in Fig.~\ref{Fig_Novelty}).
\begin{itemize}[leftmargin=*]
    \item \textbf{Identification of the targeted pruning layers (Sec.~\ref{Sec_PrimeSVT_TargetLayers}):} 
    It defines the appropriate targeted layers for pruning while avoiding an accuracy drop. To achieve this, we evaluate the impact of different rates of uniform structured pruning (i.e., \textit{channel-wise pruning}) on different sets of top-$L$ largest layers.
    \item \textbf{Prioritized compression policy (Sec.~\ref{Sec_PrimeSVT_Compress})}:
    It performs structured pruning across the targeted layers sequentially from the largest layer to the smallest one. 
    It employs a non-uniform pruning rate, which is determined through \textit{our proposed robustness metric} based on accuracy and memory. 
    \item \textbf{Compressed model selection (Sec.~\ref{Sec_PrimeSVT_Select}):}  
    It evaluates the accuracy and memory of the model candidates, then selects the one that meets both the accuracy and memory constraints.  
\end{itemize}

\begin{figure*}[t]
\centering
\includegraphics[width=\linewidth]{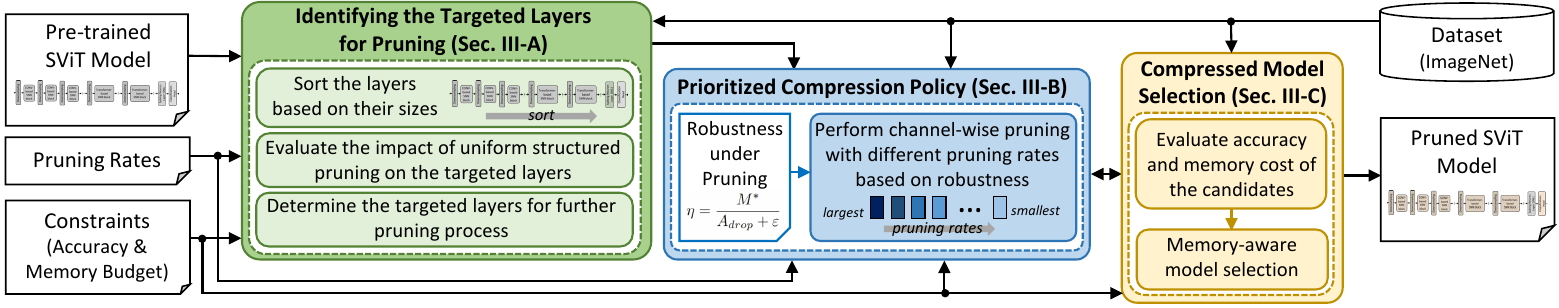}
\vspace{-0.7cm}
\caption{Our PrimeSVT framework showing its key ideas: identifying the targeted pruning layers, prioritized compression policy, and compressed model selection, which are fully automated.}
\label{Fig_PrimeSVT}
\vspace{-0.4cm}
\end{figure*}

\textbf{Key Results:}
We evaluate the PrimeSVT framework via PyTorch implementation that runs on the Nvidia GeForce RTX 4090 GPU machine. 
Experimental results show that PrimeSVT reduces 26.68\% of memory (i.e., number of parameters) through single-shot pruning, while preserving accuracy within 3\% (70.3\% without fine-tuning and 72.9\% with fine-tuning) from that of the unpruned model (73.3\%), meeting the accuracy and memory constraints.

\section{Background}
\label{Sec_Back}

\begin{figure}[t]
\centering
\includegraphics[width=\linewidth]{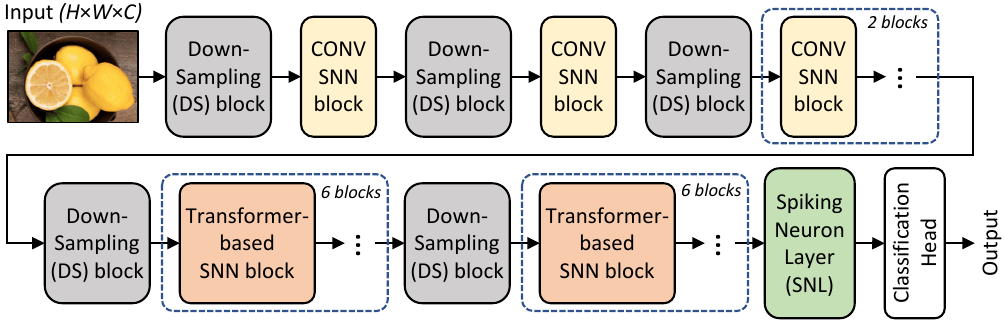}
\vspace{-0.7cm}
\caption{Overview of the SDTv2 network architecture~\cite{Ref_Yao_SpikeDrivenTransformer2_ICLR24}.}
\label{Fig_SDTv2}
\vspace{-0.3cm}
\end{figure}

An SNN model is built from network architecture, spiking neuron, neural coding, and learning rule~\cite{Ref_Maass_SNN_NeuNet97}\cite{Ref_Mozafari_SpykeTorch_FNINS19}. 
To enable the practicality of SNNs for diverse ultra-low power/energy applications, efforts for advancing SNNs span widely, encompassing software part~\cite{Ref_Putra_FSpiNN_TCAD20}\cite{Ref_Rathi_PruneQuantizeSNN_TCAD18}\cite{Ref_Minhas_Replay4NCL_DAC25, Ref_Chowdhury_TemporalPruneSNN_ECCV22, Ref_Minhas_SurveyNCL_Access25}, hardware part~\cite{Ref_Akopyan_TrueNorth_TCAD15, Ref_Davies_Loihi_MM18, Ref_Frenkel_ODIN_TBCAS19, Ref_Neckar_Braindrop_IEEE19, Ref_SynSense_DYNAP, Ref_BrainChip_Akida}, and cross-layer methods~\cite{Ref_Putra_ReSpawn_ICCAD21, Ref_Rathi_SNNsurvey_CSUR23, Ref_Putra_SoftSNN_DAC22, Ref_Putra_SpikeNAS_TAI26}.  
Currently, major SViT advancements still aim to achieve higher accuracy, such as Spikformer~\cite{Ref_Zhou_Spikformer_ICLR23}, Spike-driven Transformer (SDT)~\cite{Ref_Yao_SpikeDrivenTransformer_NeurIPS23}, and Spike-driven Transformer (SDTv2) ~\cite{Ref_Yao_SpikeDrivenTransformer2_ICLR24}.
Despite having different detailed layers, these models have similar major structural blocks, i.e., transformer, conventional (CONV), and feed-forward (FF) parts. Hence, applying a pruning method to one model makes it applicable for other models.
In this work, we consider the SDTv2~\cite{Ref_Yao_SpikeDrivenTransformer2_ICLR24} as the state-of-the-art SViT model, since it offers state-of-the-art accuracy with competitive memory requirements as compared to other designs (i.e., achieving almost 80\% accuracy on the ImageNet-1K dataset with 55.4M parameters). 
An overview of SDTv2 architecture is presented in Fig~\ref{Fig_SDTv2}, comprising different blocks, such as downsampling (DS), CONV-based SNN, and transformer-based SNN. 

\section{The PrimeSVT Framework}
\label{Sec_PrimeSVT}

This framework employs three main ideas: identification of the targeted pruning layers, prioritized compression policy, and compressed model selection (see Fig.~\ref{Fig_PrimeSVT}), as described below.

\subsection{Identifying the Targeted Pruning Layers}
\label{Sec_PrimeSVT_TargetLayers}

\subsubsection{\textbf{Strategy}}
\label{Sec_PrimeSVT_TargetLayers_Strategy}

Based on our observations in Sec.~\ref{Sec_Intro_Challenges}, different layers may have different sensitivity levels under pruning; see Fig.~\ref{Fig_CaseStudy}(b).
Therefore, identifying \textit{the initial targeted layers} for pruning is important for achieving effective memory saving and accuracy preservation. 
To do this, PrimeSVT conducts an experimental-based decision making with the following steps (pseudocode is in Alg.~\ref{Alg_InitTargetLayers}).
\begin{itemize}[leftmargin=*]
    \item Sort the layers from the largest one to the smallest one. 
    \item Define different sets of candidates for targeted pruning layers ($candSets$), e.g., 100\% (all layers), top-95\% largest layers, top-90\% largest layers, etc. 
    Here, we employ the top largest layers as they can offer higher memory saving when pruned. 
    \item Perform uniform \textit{channel-wise filter pruning} (see Sec.~\ref{Sec_PrimeSVT_TargetLayers_PruneAlg} and Alg.~\ref{Alg_CWFprune}) for each set of candidates across different pruning rates, and then evaluate their accuracy and memory requirements. 
    \item Then, we select the appropriate candidate for the initial set of targeted pruning layers ($targetSet$) via the following. 
    \begin{itemize}
        \item Calculate \textit{the average accuracy} ($\hat{A}$) for each candidate from its accuracy scores across different pre-defined pruning rates ($R$) that are still within the acceptable accuracy constraint ($const_A$), and \textit{the average memory saving} ($\hat{M}^*$).  
        \item Select the candidate that offers the highest benefit score ($B$) as the $targetSet$; see Eq.~\ref{Eq_Benefit}. 
        \item Select the highest pruning rate in $R$ that leads to acceptable accuracy for $targetSet$, since the initial pruning rate ($r_{init}$).
        \begin{equation}
          B = \hat{A} + \hat{M}^*
          \label{Eq_Benefit}
        \end{equation}
    \end{itemize}
\end{itemize}

\begin{algorithm}[h]
\caption{Selection of the initial targeted pruning layers}
\label{Alg_InitTargetLayers}
\footnotesize
\begin{algorithmic}[1]
\renewcommand{\algorithmicrequire}{\textbf{INPUT:}}
\renewcommand{\algorithmicensure}{\textbf{OUTPUT:}}
\REQUIRE \textbf{(1)} Pre-trained SViT model ($Net$), its accuracy ($Acc_0$) and memory ($Mem_0$);  
\textbf{(2)} Pre-defined sets of candidates for targeted layers ($candSets$ = $\{\text{top-100\% largest layers}, \text{top-95\% largest layers}, ...\}$); Number of pre-defined sets ($N_{Set}$); 
\textbf{(3)} Pre-defined pruning rates ($R$); Number of pre-defined rates ($N_{R}$); 
\textbf{(4)} Acceptable accuracy degradation ($const_A$);
\ENSURE Initial set of the targeted layers ($targetSet$); \\
\smallskip
\textbf{BEGIN} \\
  \textbf{Initialization}: \\
    \STATE $n$ = 0; \\
    \STATE $Acc$ = zeros($N_{Set}$,$N_{R}$); \\
    \STATE $Mem$ = zeros($N_{Set}$,$N_{R}$); \\
  \textbf{Process}: \\
    \FOR{($s$ = 1; $s<$ ($N_{Set}$+1); $s$++)}
      \FOR{($r$ = 1; $r<$ ($N_R$+1); $r$++)}
        \STATE $Net_{p}$ = channel\_wise\_prune($Net, candSets[s], R[r]$); \textcolor{teal}{// Alg.~\ref{Alg_CWFprune}}
        \STATE $Acc[s,r]$, $Mem[s,r]$ = eval($Net_{p}$); 
        \IF{($\big|Acc_0$ - $Acc[s,r]\big| \leq const_A$)}
          \STATE $n$ = $n$+1;
          \STATE $M^*[s,n]$ = mem\_saving($Mem[s,r]$, $Mem_0$); 
          \STATE $A[s,n]$ = $Acc[s,r]$;
        \ENDIF
      \ENDFOR 
      \STATE $B[s]$ = $\frac{1}{n} \sum_{i=1}^{n} A[s,i]$ + $\frac{1}{n} \sum_{i=1}^{n} M^*[s,i]$; \textcolor{teal}{// Eq.~\ref{Eq_Benefit}}
      \STATE $n$ = 0; \\
    \ENDFOR 
    \STATE $targetSet$ = select($candSets$, $\max_{s}$($B[s]$));
  \RETURN $targetSet$; \\
\textbf{END}
\end{algorithmic} 
\end{algorithm}
\setlength{\textfloatsep}{1pt}

\subsubsection{\textbf{Channel-wise Filter Pruning Algorithm}}
\label{Sec_PrimeSVT_TargetLayers_PruneAlg}

\begin{figure}[t]
\centering
\includegraphics[width=\linewidth]{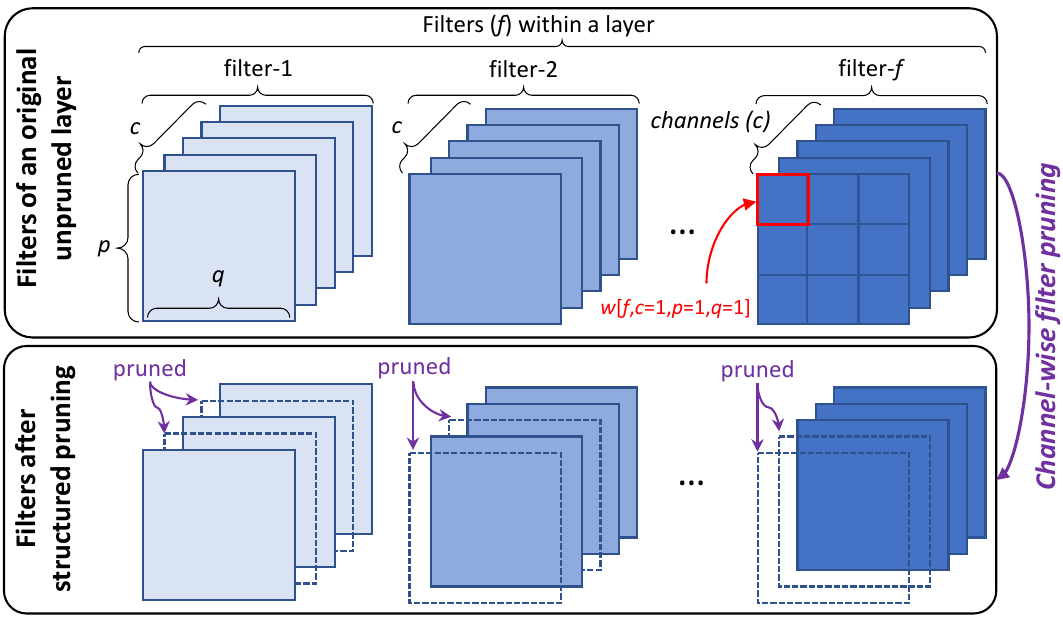}
\vspace{-0.7cm}
\caption{Conceptual illustration of the channel-wise filter pruning employed in this work.}
\label{Fig_CWPrune}
\vspace{0.4cm}
\end{figure}

Its mechanism is highlighted in Fig.~\ref{Fig_CWPrune}, whose key ideas are described below.
\begin{itemize}[leftmargin=*]
    \item We consider removing the entire channels in each filter ($f$), that have the lowest significance scores ($\sigma$) based on the pruning rate ($r$).
    Each $\sigma$ score represents the magnitude of influence made by each channel of filters on the outputs of the respective layer.
    \item The $\sigma$ score is computed using the L2-norm function in each filter channel $c$; see Eq.~\ref{Eq_L2norm}. 
    Here, $c$ denotes the channel index of filter-$f$, while $p$ and $q$ denote the row and column indices of the weight $w$, respectively.
    For point-wise convolution and linear $p=q=1$. 
    
    \begin{equation}
      \sigma = \sqrt{\sum_{p,q} (w[f,c,p,q])^2}
      \label{Eq_L2norm}
    \end{equation}
    
    \item This procedure is repeated for every weight filter across all layers. 
\end{itemize}

\begin{algorithm}[h]
\caption{Our channel-wise pruning procedure}
\label{Alg_CWFprune}
\footnotesize
\begin{algorithmic}[1]
\renewcommand{\algorithmicrequire}{\textbf{INPUT:}}
\renewcommand{\algorithmicensure}{\textbf{OUTPUT:}}
\REQUIRE \textbf{(1)} Pre-trained SViT model ($Net$) and its weights ($Net.w$); 
\textbf{(2)} A set of candidate for targeted pruning layers ($cand$); 
\textbf{(3)} Pruning ratio ($r$); \\
\ENSURE Pruned network model ($Net_p$); \\
\smallskip
\textbf{BEGIN} \\
  \textbf{Process}: \\
    \STATE Copy the pre-trained weights: $w_{tmp}$ $\leftarrow$ $Net.w$;
    \STATE $subNet$ = select($Net$, $cand$);
    \STATE Get the number of filters ($F$) and their channels ($C$) per layer in $subNet$; \\
    \smallskip
    \underline{// Perform channel-wise pruning for each layer}
    \FOR{$f = 1$ to $F$}
      \FOR{$c = 1$ to $C$}
        \smallskip
        \STATE $\sigma = \sqrt{\sum_{p,q} (w_{tmp}[f,c,p,q])^2}$; \textcolor{teal}{// Eq.~\ref{Eq_L2norm}} \\
      \ENDFOR
      \STATE $k = \max(1, \lfloor \frac{r}{100} \cdot C \rceil)$;
      \STATE $\mathcal{I}_{k} \leftarrow$ indices of $k$ channels with smallest $\sigma$;
      \FOR{each $c \in \mathcal{I}_{k}$}
        \STATE $w_{tmp}[f,c,:,:] = 0$;
      \ENDFOR
    \ENDFOR
  \STATE $Net_p.w \leftarrow w_{tmp}$;
\STATE \textbf{return} $Net_p$
\end{algorithmic}
\end{algorithm}
\setlength{\textfloatsep}{2pt}

\begin{algorithm}[t]
\caption{Prioritized compression with model selection}
\label{Alg_PCP}
\footnotesize
\begin{algorithmic}[1]
\renewcommand{\algorithmicrequire}{\textbf{INPUT:}}
\renewcommand{\algorithmicensure}{\textbf{OUTPUT:}}
\REQUIRE \textbf{(1)} Pre-trained SViT model ($Net$); and its accuracy ($Acc_0$) and memory ($Mem_0$);  
\textbf{(2)} Number of robustness groups ($N_G$ = 5); \\ 
\textbf{(3)} Max. accuracy drop ($const_A$), and min. memory saving ($const_M$);\\
\textbf{(4)}  A set of sorted blocks ($target\vec{B}$) and its robustness groups ($group\vec{B}$); \\
\ENSURE Selected pruned network model ($Net^*_p$); \\
\smallskip
\textbf{BEGIN} \\
  \textbf{Process}: \\
    \STATE $targetB$ = group\_block($targetSet$); \textcolor{teal}{// Group layers into blocks} \\ 
    \STATE $N_{block}$ = get\_Nblock($targetB$); \textcolor{teal}{// Get the number of blocks} \\
    \smallskip
    \underline{// Evaluate the robustness of each block} 
    \FOR{($b$ = 1; $b<$ ($N_{block}$+1); $b$++)}
      \STATE $Net_{p1}$ = channel\_wise\_prune($Net, targetB[:], r_{init}$); \textcolor{teal}{// Alg.~\ref{Alg_CWFprune}} \\ 
      \STATE $Net_{p2}$ = channel\_wise\_prune($Net_{p1}, targetB[b], 2r_{init}$); \textcolor{teal}{// Alg.~\ref{Alg_CWFprune}} \\ 
      \STATE $Acc_{b}$, $Mem_{b}$ = eval($Net_{p2}$); \\
      \STATE $A_{drop}[b]$ = acc\_drop($Acc_{b}$, $Acc_0$); \\
      \STATE $M^*[b]$ = mem\_saving($Mem_{b}$, $Mem_0$); \\
      \STATE $\eta[b]$ = $\frac{M^*[b]}{A_{drop}[b]+\varepsilon}$; \textcolor{teal}{// Eq.~\ref{Eq_RobustMetric}} \\
    \ENDFOR 
  \STATE $target\vec{B}$ = sort($targetB$, $\eta$[:]);  \textcolor{teal}{// Sort the blocks in descending $\eta$ order} 
  \STATE $\eta_{max}$ = $\max_{b}$($\eta[b]$);
  \\
  \smallskip
  \underline{// Group the blocks based on $\eta$} \\
  \smallskip
  \FOR{($b$ = 1; $b<$ ($N_{block}$+1); $b$++)}
      \STATE $g$ = check\_group($target\vec{B}[b]$, $\eta[b]$, $\eta_{max}$); \textcolor{teal}{// Use info from Table~\ref{Tab_RobustLevels}} \\
      \STATE $group\vec{B}[b]$ = $g$; \textcolor{teal}{// Record the group classification} \\
  \ENDFOR 
  \smallskip
  \underline{// Perform incremental prioritized pruning} 
  \smallskip
  \STATE $Net^*_p$ = \{\}; \\
  \STATE $r_{base}[:]$ = $r_{init}$; \\
  \STATE $M^*_{tmp}$ = 0; \\
  \WHILE{(1)}
    \FOR{($b$ = 1; $b<$ ($N_{block}$+1); $b$++)}
    \STATE $flag$ = 0;
    \FOR{($g$ = $group\vec{B}[b]$; $g<$ ($N_{G}$+1); $g$++)}
      \IF{($g$ == $group\vec{B}[b]$)}
        \STATE $r_{inc}[b]$ = get\_rate($R_{inc}$, $group\vec{B}[b]$); \textcolor{teal}{// $R_{inc}$$\in$\{8\%, ..., 0\%\}}
      \ENDIF
      \IF{($flag$ == 0)}
        \STATE $r_{tmp}[b]$ = $r_{base}[b]$+$r_{inc}[b]$;
        \STATE $Net_{t}$ = channel\_wise\_prune($Net$, $target\vec{B}[b]$, $r_{tmp}[b]$); 
        \STATE $Acc_{t}$, $Mem_{t}$ = eval($Net_{t}$); \\
        \STATE $Net^*_p$ = cand($Net_{t}$, $Acc_{t}$, $Mem_{t}$, $Acc_0$, $Mem_0$); \textcolor{teal}{// Alg.~\ref{Alg_Update}}
        \IF{($A_{drop} \leq const_A$)}
          \STATE $r_{base}[b]$ = $r_{tmp}[b]$;
          \STATE $flag$ = 1;
        \ELSE
          \IF{($g$ == ($N_G$-1))}
            \STATE $flag$ = 1;
          \ELSE
            \STATE $r_{inc}[b]$ = $r_{inc}[b]/2$;
          \ENDIF
        \ENDIF
      \ENDIF
    \ENDFOR 
    \ENDFOR 
    \IF{(no additional pruning $r_{inc}[:]$ across blocks)}
      \STATE \textbf{break};  
    \ENDIF
  \ENDWHILE 
  \IF{($Net^*_p[1]$ is not $\varnothing$)}
    \STATE \textbf{return} $Net^*_p[1]$; \textcolor{teal}{// Meeting both constraints ($const_A$ and $const_M$)}
  \ELSE
      \IF{($Net^*_p[2]$ is not $\varnothing$) \AND ($Net^*_p[3]$ is not $ \varnothing$)}
        \STATE \textbf{return} $Net^*_p[2:3]$; \textcolor{teal}{// Alternatives if all constraints are not met at once}
      \ENDIF
      \IF{($Net^*_p[2]$ is not $\varnothing$) \AND ($Net^*_p[3]$ is $ \varnothing$)}
        \STATE \textbf{return} $Net^*_p[2]$; \textcolor{teal}{// Meeting only accuracy constraint ($const_A$)}
      \ENDIF
      \IF{($Net^*_p[2]$ is $\varnothing$) \AND ($Net^*_p[3]$ is not $\varnothing$)}
        \STATE \textbf{return} $Net^*_p[3]$; \textcolor{teal}{// Meeting only memory constraint ($const_M$)}
      \ENDIF 
\ENDIF 
\textbf{END}
\end{algorithmic} 
\end{algorithm}
\setlength{\textfloatsep}{4pt}

\subsection{Prioritized Compression Policy}
\label{Sec_PrimeSVT_Compress}
\vspace{-0.2cm}

This policy aims to perform structured pruning to the targeted layers (determined in Sec.~\ref{Sec_PrimeSVT_TargetLayers_Strategy}) by leveraging the channel-wise filter pruning (determined in Sec.~\ref{Sec_PrimeSVT_TargetLayers_PruneAlg}) across layers to maximize memory saving and preserve high accuracy.  
Its key idea is \textit{to compress different layers with different pruning rates based on their robustness under pruning}. 
It is realized via the following steps; see the pseudocode in Alg.~\ref{Alg_PCP}.

\begin{itemize}[leftmargin=*]
    \item We group the targeted layers within $targetSet$ into blocks ($targetB$) based on the hierarchy of network architecture.
    For instance, SDTv2 blocks are shown in Fig.~\ref{Fig_SDTv2}. 
    \item We investigate the robustness of each block to identify how to safely prune the block for saving memory while preserving high accuracy. 
    It is done through the steps below. 
    \begin{itemize}
        \item Compress the investigated block with pruning rate of 2$r_{init}$ while compressing the other blocks with pruning rate of $r_{init}$ ($r_{init}$ is obtained from Sec.~\ref{Sec_PrimeSVT_TargetLayers_Strategy}), then evaluate the accuracy reduction ($A_{drop}$) and memory saving ($M^{*}$). 
        \item Compute the robustness level of the investigated block using our novel \textit{robustness-under-pruning metric} ($\eta$) stated as Eq.~\ref{Eq_RobustMetric}. 
        
        \begin{equation}
            \eta = \frac{M^{*}}{A_{drop} + \varepsilon} \;\;\;\; \text{with} \;\;\;\; \varepsilon = 0.001
            \label{Eq_RobustMetric}
        \end{equation} 
        In this manner, robustness ($\eta$) becomes high when memory saving $M^*$ is high and accuracy degradation $A_{drop}$ is low.  
        Here, $\varepsilon$ is a constant to avoid division by 0.
    \end{itemize} 
    \item Then, we rank all targeted blocks based on their robustness, and define their suitable incremental pruning rate ($r_{inc}$) values to be added on top of the initial one $r_{init}$. 
    Hence, we potentially apply different $r_{inc}$ values for different robustness levels to carefully compress the respective blocks.
    \item To determine $r_{inc}$ values, we need to quantitatively compare the relative robustness levels across blocks.
    To do this, we consider 5 robustness groups based on the maximum $\eta$ value across all blocks ($\eta_{max}$), robustness range ($\theta$), and robustness level of the targeted block ($\eta_b$); see Table~\ref{Tab_RobustLevels}. 
    Then, we assign $r_{inc}$ with a pre-defined value based on the respective group.
    \item We perform an incremental prioritized pruning until it finds the most suitable solution, whose mechanism is provided in Alg.~\ref{Alg_PCP}: lines 15-47. 
    If $r_{inc}$ causes significant accuracy degradation beyond acceptable accuracy constraint ($const_A$), then we halve $r_{inc}$ to ensure a more careful pruning (i.e., \textit{our reduction policy for $r_{inc}$}); see Alg.~\ref{Alg_PCP}: lines 29-36. 
\end{itemize}

\begin{table}[t]
\caption{The robustness groups, with the default value of $\theta$=1.}
\small
\centering
\begin{tabular}{|c|c|c|}
\hline
  \textbf{Group} & \textbf{Categorization}  & \textbf{Action} \\ \hline \hline
  Group-1 & $\eta_b \geq \eta_{max}-\theta$ & Apply $r_{inc} = 8\%$  \\  \hline
  Group-2 & $\eta_{max}-\theta > \eta_b \geq \eta_{max}-2\theta$ & Apply $r_{inc} = 4\%$  \\  \hline
  Group-3 & $\eta_{max}-2\theta > \eta_b \geq \eta_{max}-3\theta$ & Apply $r_{inc} = 2\%$  \\  \hline
  Group-4 & $\eta_{max}-3\theta > \eta_b \geq \eta_{max}-4\theta$ & Apply $r_{inc} = 1\%$  \\  \hline
  Group-5 & $\eta_{max}-4\theta > \eta_b \geq \eta_{max}-5\theta$ & Apply $r_{inc} = 0\%$  \\  \hline
\end{tabular}
\label{Tab_RobustLevels}
\end{table}

\subsection{Compressed Model Selection}
\label{Sec_PrimeSVT_Select}

It evaluates the accuracy and memory of the model candidates, and then selects the one that meets accuracy and memory constraints ($const_A$ and $const_M$, respectively). 
Here, there are several possible outcomes as multiple constraints may not be satisfied at once. 
Hence, PrimeSVT also considers different possible outcomes during its \textit{search-and-update process} and \textit{model selection process} with the following mechanisms.

\smallskip
\subsubsection{\textbf{Search-and-update process}} 
Its mechanism is presented in Alg.~\ref{Alg_PCP}: line 28 and Alg.~\ref{Alg_Update}, and described below.
\begin{itemize}[leftmargin=*]
    \item It calculates the accuracy drop $A_{drop}$ and the memory saving $M^*$ of the given pruned model $Net_t$. 
    \item If the candidate only meets $const_A$, then it is recorded as $best\_acc$. 
    If the candidate only meets $const_M$, then it is recorded as $best\_mem$.
    Meanwhile, if the candidate meets both $const_A$ and $const_M$, then it is recorded as $best\_both$.
    \item Then, it continues the search process, and updates $best\_acc$, $best\_mem$, or $best\_both$ when the model candidate passes the criteria, until no additional incremental pruning $r_{inc}$. 
\end{itemize}

\subsubsection{\textbf{Model selection process}}
If the search-and-update process is finished, then PrimeSVT selects and returns the pruned model through the following criteria.
\begin{itemize}[leftmargin=*]
    \item If the search-and-update process finds a pruning configuration that can meet both $const_A$ and $const_M$, then we return the corresponding model $Net^*_p[1]$ (see Alg.~\ref{Alg_PCP}: lines 39-40).
    \item If no pruning configuration meets both $const_A$ and $const_M$, then there are two possible conditions.
    \begin{itemize}
        \item If there are pruned models stored in $best\_acc$ ($Net^*_p[2]$) and $best\_mem$ ($Net^*_p[3]$) during the search-and-update process, they will be returned as alternatives; see Alg.~\ref{Alg_PCP}: lines 42-43. 
        \item If there is only one pruned model stored in either $best\_acc$ or $best\_mem$ during the search-and-update process, then this model is returned as the alternative; see Alg.~\ref{Alg_PCP}: lines 44-47.  
    \end{itemize}
\end{itemize}

\begin{algorithm}[t]
\caption{Search-and-update process for model candidates}
\label{Alg_Update}
\footnotesize
\begin{algorithmic}[1]
\renewcommand{\algorithmicrequire}{\textbf{INPUT:}}
\renewcommand{\algorithmicensure}{\textbf{OUTPUT:}}
\REQUIRE Model candidate ($Net_t$), its accuracy ($Acc$) and memory ($Mem$); \\
\ENSURE Updated sets of candidates ($best\_acc$, $best\_mem$, $best\_both$); \\
\smallskip
\textbf{BEGIN} \\
\STATE $A_{drop}$ = acc\_drop($Acc$, $Acc_0$);
\STATE $M^*$ = mem\_saving($Mem$, $Mem_0$);

\smallskip
\textbf{Process}:
\IF{($A_{drop} \leq const_A$)}
  \IF{($best\_acc$ is $\varnothing$) \OR ($M^* > best\_acc.M^*$)}
    \STATE $best\_acc = \{Net_t, Acc, A_{drop}, M^*\}$;
  \ENDIF
\ENDIF

\IF{($M^* \geq const_M$)}
  \IF{($best\_mem$ is $\varnothing$) \OR ($A_{drop} < best\_mem.A_{drop}$)}
    \STATE $best\_mem = \{Net_t, Acc, A_{drop}, M^*\}$;
  \ENDIF
\ENDIF

\IF{($A_{drop} \leq const_A$) \AND ($M^* \geq const_M$)}
  \IF{($best\_both$ is $\varnothing$) \OR ($M^* > best\_both.M^*$)}
    \STATE $best\_both = \{Net_t, Acc, A_{drop}, M^*\}$;
  \ENDIF
\ENDIF

\STATE $Net^*_p[2]$ = $best\_acc$; 
\STATE $Net^*_p[3]$ = $best\_mem$;
\STATE $Net^*_p[1]$ = $best\_both$;
    
\STATE \textbf{return} $Net^*_p$; \\
\textbf{END}
\end{algorithmic}
\end{algorithm}
\setlength{\textfloatsep}{4pt}

\vspace{-0.2cm}
\section{Evaluation Methodology}
\label{Sec_EvalMethod}

We implement the PrimeSVT framework using PyTorch and based on the SpikingJelly~\cite{Ref_Fang_SpikingJelly_SciAdv23}. 
Then, we run it on a single Nvidia RTX 4090 GPU device to ensure consistent hardware conditions. 
For the baseline model (i.e., original/unpruned one), we consider using the state-of-the-art SDTv2~\cite{Ref_Yao_SpikeDrivenTransformer2_ICLR24}. 
We employ its publicly available codes, then reproduce its training, validation, and testing while considering its default hyperparameter settings on the ImageNet-1K dataset~\cite{Ref_Deng_ImageNet_CVPR09}. 
Under these settings, we achieve 73.3\% top-1 accuracy for the baseline SDTv2. 
For the constraints, we consider 25\% of the minimum memory saving ($const_M)$ with 3\% maximum accepted accuracy degradation ($const_A)$.
Our experiments evaluate several metrics, such as top-1 accuracy, number of weight parameters (memory cost), computational cost, and energy consumption.
For energy consumption, we leverage the analytical model of energy consumption for SNNs~\cite{Ref_Lemaire_SNNenergyModel_ICONIP22} and the parameter values for a single operation (e.g., an addition and a memory access) under 45nm CMOS technology based on~\cite{Ref_Jouppi_10Lessons_ISCA21} to ensure reproducible energy estimation. 

\section{Results and Discussion}
\label{Sec_Results}

\subsection{Reducing Memory and Preserving Accuracy}
\label{Sec_Results_Memory}

Experimental results for memory cost and accuracy are shown in Table~\ref{Tab_Comparison}.
They show that our PrimeSVT effectively compresses the baseline SDTv2 by 26.68\% (i.e., 40.6M parameters) through single-shot pruning, while preserving high accuracy within 3\% acceptable degradation (i.e., 70.32\% accuracy without fine-tuning and 72.9\% accuracy with fine-tuning, approaching the baseline accuracy). 
%
\begin{table}[h]
\caption{Comparison of the state-of-the-art SVIT models and their pruning methods on the ImageNet-1K. 
Note: *) Accuracy scores for SDTv2 and PrimeSVT are from our experiments; \\ $\dagger$) Accuracy score is obtained from fine-tuning after pruning.}
\label{Tab_Comparison}
\scriptsize
\centering
\begin{tabular}{|c|c|c|c|c|c|}
\hline
\textbf{Category} & \textbf{Method} & \begin{tabular}[c]{@{}c@{}} \textbf{Reference} \\ \textbf{Network} \end{tabular} & \textbf{\begin{tabular}[c]{@{}c@{}} \textbf{\# Weights} \\ \textbf{[M]} \end{tabular}} & \begin{tabular}[c]{@{}c@{}} \textbf{Time-} \\ \textbf{step} \end{tabular} & \textbf{\begin{tabular}[c]{@{}c@{}} \textbf{Top-1} \\ \textbf{Acc. [\%]} \end{tabular}} \\
\hline
\hline
\multicolumn{2}{|c|}{} & Spikformer~\cite{Ref_Zhou_Spikformer_ICLR23} & 66.3 & 4 & 74.8$\;\;$ \\
\cline{3-6}
\multicolumn{2}{|c|}{Baseline} & SDT~\cite{Ref_Yao_SpikeDrivenTransformer_NeurIPS23} & 66.3 & 4 & 76.3$\;\;$ \\
\cline{3-6}
\multicolumn{2}{|c|}{} & SDTv2~\cite{Ref_Yao_SpikeDrivenTransformer2_ICLR24} & 55.4 & 4 & 73.3$^*$ \\
\hline
Token & STSFP~\cite{Ref_Zhou_STSFP_TCDS25}
& SDTv2~\cite{Ref_Yao_SpikeDrivenTransformer2_ICLR24} & 55.4 & 4 & 83.1$^\dagger$ \\
\cline{2-6}
Pruning & STATA~\cite{Ref_Zhuge_TokenSparse_ICML24} & Spikformer~\cite{Ref_Zhou_Spikformer_ICLR23} & 66.3 & 4 & 74.0$\;\;$ \\
\hline
\begin{tabular}[c]{@{}c@{}} \textbf{Weight} \\ \textbf{Pruning} \end{tabular} & \begin{tabular}[c]{@{}c@{}} \textbf{PrimeSVT} \\ \textbf{(Ours)}\end{tabular}
& \textbf{SDTv2}~\cite{Ref_Yao_SpikeDrivenTransformer2_ICLR24} & \textbf{40.6} & \textbf{4} & \begin{tabular}[c]{@{}c@{}} \textbf{70.32}$^*$ \\ \textbf{72.9}$^{*\dagger}$\end{tabular}  \\
\hline
\end{tabular}
\end{table}
\setlength{\textfloatsep}{6pt}

These significant memory saving and high accuracy preservation come from the following design choices in the PrimeSVT.
\begin{itemize}[leftmargin=*]
    \item It prioritizes to prune larger layers than smaller ones during the identification of targeted layers, thereby potentially facilitating higher pruning rates with acceptable accuracy. 
    \item It compresses different targeted layers with different pruning rates based on their robustness levels, hence allowing higher pruning rates to be applied to layers with higher robustness while maintaining high accuracy. 
    \item It carefully applies incremental pruning rates $r_{inc}$ based on the pre-defined values and their reduction policy when pre-defined ones degrade accuracy beyond the constraint $const_A$. 
\end{itemize}

\vspace{-0.2cm}
\subsection{Reducing Computation and Energy Costs}
\label{Sec_Results_ComputeEnergy}

\begin{figure}[h]
\centering
\includegraphics[width=\linewidth]{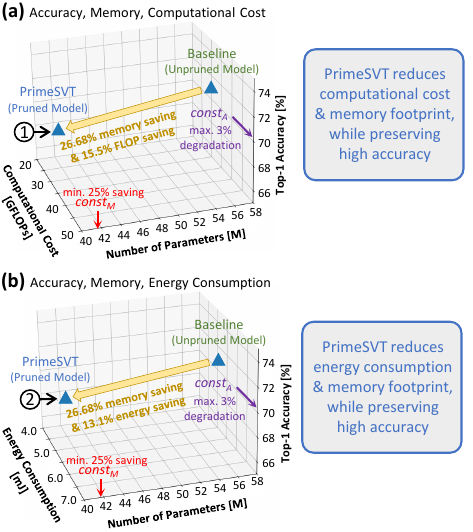}
\vspace{-0.6cm}
\caption{Results for $const_A$= 3\% and $const_M$= 25\%: \textbf{(a)} accuracy, memory, and computational cost; and \textbf{(b)} accuracy, memory, and energy consumption.}
\label{Fig_Results_Main}
\vspace{-0.3cm}
\end{figure}

Experimental results for computational cost are presented in Fig.~\ref{Fig_Results_Main}(a). 
These results show that our PrimeSVT effectively reduces the floating-point operations (FLOPs) by 15.5\% in the pruned model with 35.5 GFLOPs, as compared to the baseline model with 42 GFLOPs; as indicated by \circled{1}.
This reduction comes from the reduction of weight parameters in the pruned model.
Furthermore, we also evaluate and compare the energy consumption between the baseline model with the pruned model, and the experimental results are presented in Fig.~\ref{Fig_Results_Main}(b). 
These results show that our PrimeSVT effectively reduces energy consumption by 13.1\% compared to the baseline model due to its significant reduction in the number of weight parameters and computational cost, as indicated by \circled{2}. 

\begin{figure*}[t]
\centering
\includegraphics[width=\linewidth]{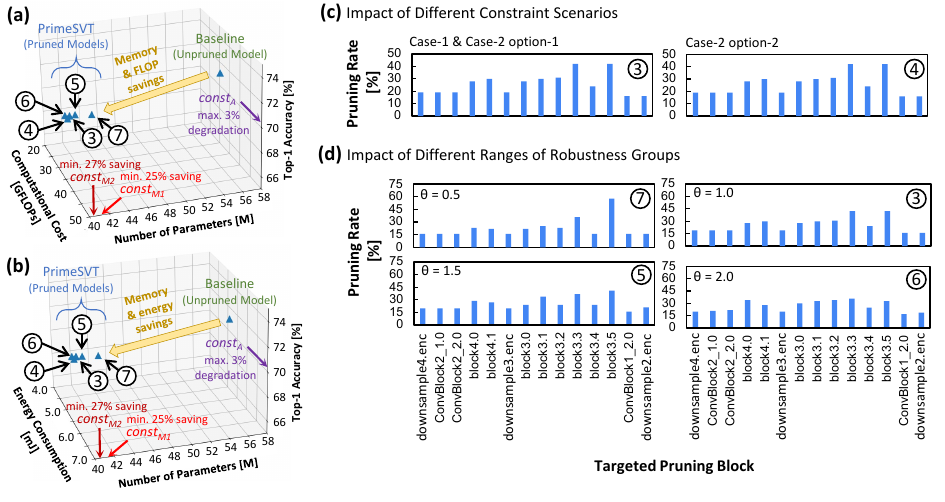}
\vspace{-0.8cm}
\caption{Results for ablation study: \textbf{(a)} accuracy, memory, and computational cost; \textbf{(b)} accuracy, memory, and energy consumption; \textbf{(c)} pruning configurations considering different constraints; and \textbf{(d)} pruning configurations considering different ranges of robustness groups $\theta$.}
\label{Fig_Results_Ablation}
\vspace{-0.3cm}
\end{figure*}

\subsection{Ablation Study and Further Discussion}
\label{Sec_Results_Ablation}

\subsubsection{\textbf{Impact of Different Constraint Scenarios}}
\label{Sec_Results_Ablation_Constraints}

Multiple constraints may not be met at once due to conflicting objectives, such as high accuracy and small memory cost.
Hence, PrimeSVT provides alternative pruned models when both accuracy constraint ($const_A$) and memory constraint ($const_M$) are not met at once.   
To show this, we conduct experiments with different cases of constraints below. 
\begin{itemize}[leftmargin=*]
    \item Case-1: For $const_A$ = 3\% and $const_M$ = 25\%, PrimeSVT finds a pruned model that achieves 70.32\% accuracy, 26.68\% memory saving, 15.5\% reduction of FLOPs, and 13.1\% energy saving, meeting accuracy and memory constraints at once; see \circled{3} in Fig.~\ref{Fig_Results_Ablation}. 
    \item Case-2: For $const_A$ = 3\% and $const_M$ = 27\%, PrimeSVT does not find a pruned model that can satisfy both constraints at once. Hence, PrimeSVT provides two alternatives as described below. 
    \begin{enumerate}
        \item A pruned model that only meets accuracy constraint with 70.32\% accuracy, 26.68\% memory saving, 15.5\% reduction of FLOPs, and 13.1\% energy saving; see \circled{3} in Fig.~\ref{Fig_Results_Ablation}.
        \item A pruned model that only meets memory constraint with 70.1\% accuracy, 27\% memory saving, 15.7\% reduction of FLOPs, and 13.3\% energy saving; see \circled{4} in Fig.~\ref{Fig_Results_Ablation}. 
    \end{enumerate}
\end{itemize}
For these study cases, PrimeSVT takes about 11.5 hours on average to find the solution. 
These results demonstrate the capability of PrimeSVT to handle different scenarios of constraints, which comes from its model selection strategy that already covers different possible outcomes from the search-and-update process. 

\subsubsection{\textbf{Impact of Different Ranges of Robustness Groups ($\theta$)}}
\label{Sec_Results_Ablation_Robust}

We also evaluate the impact of different ranges of robustness groups through experiments, by considering $\theta$$\in$$\{0.5, 1, 1.5, 2\}$ with $const_A$ = 3\% and $const_M$ = 25\%. 
The experimental results are shown in Fig.~\ref{Fig_Results_Ablation}. 
Here, we observe that different $\theta$ values may lead to different solutions. 
For instance, $\theta$ values of 1, 1.5, and 2 lead PrimeSVT to find pruned models that meet both $const_A$ and $const_M$; see Fig.~\ref{Fig_Results_Ablation} labels \circled{3}, \circled{5}, and \circled{6}, respectively. 
Meanwhile, $\theta$ value of 0.5 leads PrimeSVT to find a model that only meets accuracy constraint $const_A$; see \circled{7} in Fig.~\ref{Fig_Results_Ablation}. 
The reason is that a larger $\theta$ value typically leads to a larger number of layers in the same robustness group. 
This condition increases the possibility of applying a high incremental pruning rate $r_{inc}$ to a larger number of layers, and thereby finding better pruned models that meet the memory constraint $const_M$. 

\section{Conclusion}
\label{Sec_Conclude}

In this paper, we propose the novel PrimeSVT framework to perform automated single-shot structured pruning on the pre-trained SViT models. 
For 25\% memory saving and 3\% accuracy constraints, our PrimeSVT successfully finds the solution with 26.68\% memory reduction and accuracy within 3\% from the unpruned SViT model, thereby meeting the given accuracy and memory constraints with a scalable pruning approach.
These show that our PrimeSVT successfully extends the efforts in enabling automation of SViT optimization for embedded AI systems.

\section*{Acknowledgment}
This work was partially supported by the NYUAD Center for Interacting Urban Networks (CITIES), funded by Tamkeen under the NYUAD Research Institute Award CG001.

\bibliographystyle{IEEEtran}
\bibliography{bibliography}

\end{document}